%% file: Arxiv.tex
\documentclass{article}

\usepackage{PRIMEarxiv}

\usepackage[utf8]{inputenc} 
\usepackage[T1]{fontenc}    
\usepackage{hyperref}       
\usepackage{url}            
\usepackage{booktabs}       
\usepackage{amsfonts}       
\usepackage{nicefrac}       
\usepackage{microtype}      
\usepackage{lipsum}
\usepackage{fancyhdr}       
\usepackage{graphicx}       
\usepackage{hyperref}
\graphicspath{{media/}}     

\title{Extracting Built Environment Features for Planning Research with Computer Vision: A Review and Discussion of State-of-the-Art Approaches
}

\author{
  Meiqing Li*\\
  University of California, Berkeley \\
  \href{mailto:meiqing@berkeley.edu}{\texttt{meiqing@berkeley.edu}} \\
   \And
  Hao Sheng\thanks{These authors contributed equally.} \\
  Stanford University \\
  \href{mailto:haosheng@cs.stanford.edu}{\texttt{haosheng@cs.stanford.edu}}\\
}

\begin{document}
\maketitle


\section{Introduction}
As planning research develops in response to a more complex policy context, better representation of the built environment (or physical urban system) at multiple scales becomes as important as a comprehensive and inclusive representation of decision-makers and processes \cite{miller_viewpoint_2018}.  The emergence of novel data sources (e.g., OpenStreetMap, Google Street View, and satellite imagery) allows researchers to extract higher quality measures of the built environment, addressing some of the limitations in previous research regarding spatial-temporal data quality and consistency \cite{boeing_osmnx_2017, shen_streetvizor_2018, ye_measuring_2019}.

Computer science technologies, including machine learning and artificial intelligence, especially the field of GeoAI \cite{janowicz_geoai_2020}, present opportunities to advance traditional data collection techniques by extracting built environment features beyond regional scales and network resolution. This paper provides a synthesis of the literature that connects state-of-the-art computer vision technologies and proposes a framework for their potential applications to urban planning research. It is hoped that the methodological framework proposed by this review can offer insights into what can potentially make meaningful contributions by collaboration between the fields of urban planning and computer science, such as planetary-scale empirical research on the built environment and human behavior to inform sustainable urban planning and design.  

\section{Method}
We primarily reviewed two sets of literature across planning and computer science that focuses on built environment measures and deep learning image processing methods, respectively. 

Specifically, for planning literature, we searched three terms, “land use”, “built environment”, and “urban design” among articles in four general planning journals, \textit{Journal of American Planning Association, Journal of Planning Education and Research, Journal of Planning Literature, and Urban Studies} since 2010. For this review, we only include the general planning journals but not specialized ones, as the topics represent the general interests to a broader planning academic and professional community.  After reviewing the abstracts, we identified 57 articles with built environment measures of interest. 

For computer science literature, we used the combination of one of “compute vision”, “image” or “CV” and one of “urban”, “city”, “land use”, “built environment”, “street view” or “satellite” to filter the literature from the proceedings (2010-2020) of \textit{Conference on Computer Vision and Pattern Recognition (CVPR), Conference on Neural Information Processing Systems (NeurIPS), European Conference on Computer Vision (ECCV), Association for the Advancement of Artificial Intelligence Conference (AAAI), International Conference on Computer Vision (ICCV), International Joint Conference on Artificial Intelligence (IJCAI), Special Interest Group on Knowledge Discovery and Data Mining Conference (KDD)}, and preprint articles on \textit{arXiv}. After reviewing the abstracts, we identified 41 articles with built environment measures of interest.

\section{Results}
As is evident by the reviewed literature, we find the fields of urban planning and computer science, particularly computer vision (CV), share some common areas of interest regarding built environment measures (e.g., land use, infrastructure, amenities). Nevertheless, the approach and output differ in terms of data sources, geographic coverage, scale, and resolution. While there are numerous papers on the built environment in planning literature, empirical studies often work with the trade-off between sample size and spatial resolution. Among all articles reviewed, we find that studies on neighborhood-level design characteristics are limited to a few cities or metropolitan areas, while measures for national-scale studies never go beyond 5-D variables \cite{ewing_travel_2010} (Figure \ref{fig:fig-1}, Figure \ref{fig:fig-3}). 

\input{Figures/fig-1}
\input{Figures/fig-3}
\input{Figures/fig-2}

In computer vision literature, built environment measures used to be treated as one of the evaluation tasks to validate the effectiveness of computer vision algorithms (e.g., the precision and recall improvements of a neural architecture for image classification). However, more methods are being proposed in recent years to directly extract built environment features (see Figure \ref{fig:fig-2}). For example, with more geo-tagged benchmark datasets, \cite{milojevic-dupont_learning_2020, yu_deepsolar_2018, zhou_deepwind_2019} claim their novelty almost exclusively through the new measures (e.g., solar panel installation rate, distribution grid network, etc.).  

In this section, we first discuss the common measures of built environment features across both fields, the gaps and limitations in both streams of literature, as well as the opportunities, to close the gap by incorporating AI/CV algorithms into built environment feature extraction. Then we introduce the evolution, workflow, and output of convolutional neural networks (CNN) that centers CV research in recent years. Finally, we propose a general framework for applying deep-learning-based CV algorithms for different types of planning research problems. As follows, we categorize three task-based (i.e., image classification, image segmentation, and object detection) and one task-free (i.e., representation learning) image processing methods with examples of opportunities and challenges to facilitate built environment feature extraction tasks for planning research. While the applications of each method are certainly not limited to the examples we present here, they demonstrate some typical use cases.

It is worth noticing each of the following methods requires a geo-tagged image dataset of consistent quality (in terms of the resolution and perspective, etc.) and a significant amount (usually at least at the level of thousands).  Satellite images and street view images are the two major types of data used in the computer vision literature we found. Moderate Resolution Imaging Spectroradiometer (MODIS) imagery, with a resolution of 250-1000m), Landsat 8 (with a resolution of 30m), Sentinel 1/2 (with a resolution of 10-20m), National Agriculture Imagery Program (NAIP) imagery, with a resolution of 1-2m, only covering the U.S.) are the most popular public satellite products used. Some literature (e.g., \cite{rudner_multimathbf3net_2018}) also uses proprietary high-resolution images such as DigitalGlobe (0.3-10m). On the street view side, \cite{milojevic-dupont_learning_2020, wang_urban2vec_2020} use the Google Street View Static API, which provides the most recent street view panorama closest to the request locations. We also found works that combined the two data sources through their coordinates (e.g., \cite{wojna_holistic_2020}).

\subsection{Image Classification}
Image classification accepts given input images and produces a predicted classification out of many predefined categories. It has been used to identify whether an object is present or not \cite{albert_using_2017, sheng_ognet_2020, zhou_deepwind_2019}, determine whether changes have happened between two images
\cite{revaud_did_2019, boriah_land_2008, irvin_forestnet_2020}, or recognize certain patterns be-tween entities \cite{wojna_holistic_2020}.

Typically built environment measures in planning literature that falls into this category include the presence of certain types of amenities or infrastructure that associated with travel behavior \cite{cao_associations_2019}, energy consumption \cite{kaza_land_2014, stoker_pedestrian_2015}, changes in land use or built environment for longitudinal policy evaluation (e.g., \cite{guhathakurta_residential_2010, he_land_2014}, and physical typologies of housing or human settlement (e.g., \cite{ko_urban_2013}).

\subsection{Image Segmentation}
Image segmentation can be viewed as a per-pixel image classification task.  Rather than giving classification results at the image-level, it outputs a segmentation map where each pixel is labeled in one of many predefined categories \cite{senlet_segmentation_2012, yu_deepsolar_2018, rudner_multimathbf3net_2018}.

This method is suitable for obtaining quantitative neighborhood/street-level built environment measures, such as sidewalk width road space or land-use area measurement (e.g., \cite{mehta_revisiting_2018, dill_how_2014}).

\subsection{Object Detection}
Object detection aims to detect instances of semantic objects in an image. The outputs of object detection algorithms are usually a series of bounding boxes for each predicted instance. Despite the ability to capture and localize objects, it does not give accurate boundary information compared to image segmentation \cite{couture_towards_2020, seo_self-supervised_2020}. 

Object detection accurate measure of the location of certain neighborhood amenity or street design elements at a human scale (e.g., \cite{garde_form-based_2018}, for example, locating sidewalk or crosswalks from street view imagery. It is also well-suited for measuring the dynamic changes in locations across time.

\subsection{Representation Learning}
Representation learning generates a high-dimensional embedding (usually a vector of 50-200 dimensions) for each input image. This high-dimensional vector can, in turn, feed into another machine learning model that outputs specific characteristics of the environment \cite{jean_tile2vec_2018, yao_representing_2018, wang_urban2vec_2020}. 

Abstract though, this method could potentially extract comprehensive proxies for certain qualities of a neighborhood region, especially through some quality measurement such as walk score, visual quality, etc. \cite{alcorn_bike-sharing_2019, smit_influence_2011}. 

\section{Conclusion and Discussion}
In the concluding section, we summarize the opportunities and challenges of applying computer vision algorithms to extract built environment features revealed by this interdisciplinary synthesis. One of the major challenges for large-scale deployment of CV models is the cost of obtaining and processing data, including the monetary cost of accessing images, labor cost for annotating images, and computational cost for training large scale models. It requires a thoughtful evaluation of a cost-benefit analysis of this automatic process, as well as a trade-off between coverage and accuracy. Notably, the emergence of deep learning in CV comes in parallel with the increasing accessibility of at-scale image datasets, reinforcing the enhancement of models and datasets in a virtuous circle. Another big challenge is the gaps between planning knowledge and state-of-the-art computer science techniques, which require interdisciplinary collaboration to integrate planning theories or empirical knowledge into every stage of the deployment pipeline. For large-scale deployment, it is also worthwhile to examine the different implications of universal output for diverse contexts.

\bibliographystyle{unsrt}  
\bibliography{references}  

\end{document}

%% file: Figures/fig-1.tex
\begin{figure}[!ht]
    \centering
    \includegraphics[width=0.8\linewidth]{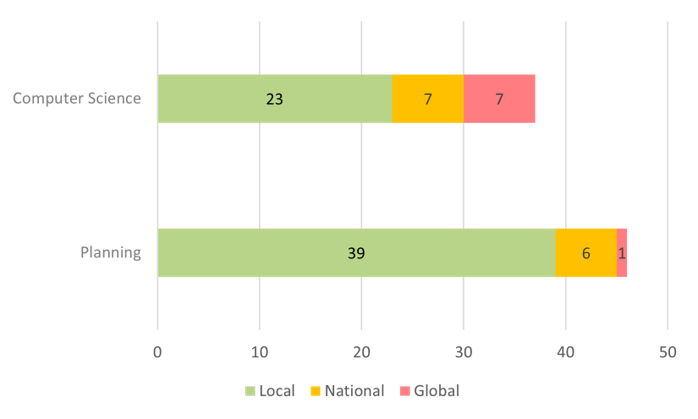}
    \caption{Distribution of geographical coverage of both urban planning and computer science literature. Studies that did not report the coverage are omitted.}
    \label{fig:fig-1}
\end{figure}

%% file: Figures/fig-3.tex
\begin{figure}[!ht]
    \centering
    \includegraphics[width=0.8\linewidth]{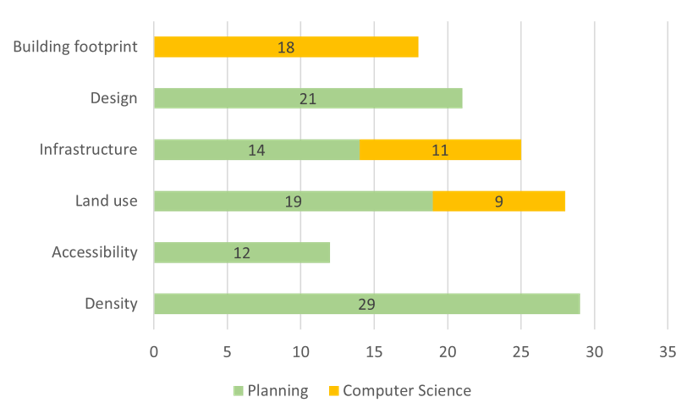}
    \caption{Built environment measures in both urban planning and computer science literature.}
    \label{fig:fig-3}
\end{figure}

%% file: Figures/fig-2.tex
\begin{figure}[!ht]
    \centering
    \includegraphics[width=0.8\linewidth]{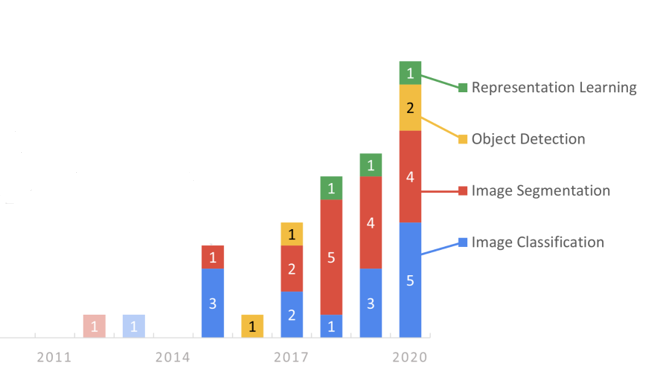}
    \caption{The number of publications in major computer science conferences that provide built environment measures.}
    \label{fig:fig-2}
\end{figure}